\documentclass[letterpaper, 10 pt, conference]{ieeeconf} 

\IEEEoverridecommandlockouts 

\usepackage{url}
\usepackage{epstopdf}
\usepackage{multirow}
\usepackage{booktabs}
\usepackage{graphicx} 
\usepackage{bm}
\usepackage{subfigure} 
\usepackage{amsfonts,amssymb} 
\usepackage[linesnumbered,boxed,ruled,commentsnumbered]{algorithm2e}
\usepackage{amsmath} 

\graphicspath{{figures/}}

\title{\LARGE \bf
Real-time Trajectory Generation for Quadrotors using\\ B-spline based Non-uniform Kinodynamic Search
}

\author{Lvbang Tang, Hesheng Wang, Peng Li and Yong Wang 
\thanks{
*This work was supported in part by the Natural Science Foundation of China under Grant U1613218, 61722309 and the Beijing Municipal Commission of Science and Technology, Intelligent Manufacturing and Robot Cultivatio under Grant Z181100003118011. Corresponding Author: Hesheng Wang.}
\thanks{Lvbang Tang and Hesheng Wang are with Department of Automation, Shanghai Jiao Tong University, Key Laboratory of System Control and Information Processing, Ministry of Education of China (email: \{tlb1768, wanghesheng\}@sjtu.edu.cn). Peng Li is with School of Mechanical engineering and automation, Harbin institute of technology (Shenzhen). Yong Wang is with Beijing Institute of Control Engineering. }
}

\begin{document}

\maketitle
\thispagestyle{empty}
\pagestyle{empty}

\begin{abstract}

In this paper, we propose a time-efficient approach to generate safe, smooth and dynamically feasible trajectories for quadrotors in the obstacle-cluttered environment. By using the uniform B-spline to represent trajectories, we transform the trajectory planning to a graph-search problem of B-spline control points in discretized space. Highly strict convex hull property of B-spline is derived to guarantee the dynamical feasibility of the entire trajectory. A novel non-uniform kinodynamic search strategy is adopted, and the step length is dynamically adjusted during the search process according to the Euclidean signed distance field (ESDF), making the trajectory achieve reasonable time-allocation and be away from obstacles. Non-static initial and goal states are allowed, therefore it can be used for online local replanning as well as global planning. Extensive simulation and hardware experiments show that our method achieves higher performance compared with the state-of-the-art method.

\end{abstract}

\section{Introduction}

Many practical applications of quadrotors require safe navigation in cluttered environments, such as search, rescue, and exploration. Motion planning is the foundation of safe navigation, which generates trajectories that are followed by the motion controller.  

Safety, smoothness and good time-allocation are important aspects of evaluating the quality of the trajectory. Smooth and dynamically feasible trajectory allows it to be well tracked by the quadrotor without significant control errors. A trajectory that maintains a safe distance from obstacles as much as possible is not only conducive to avoiding unexpected collision, but also facilitates observation of the environment due to less obstruction by obstacles. Time-allocation, also called time parameterization, affects the trajectory velocity and acceleration in different segments, which in turn affects the safety and efficiency of the execution of the trajectory. And it is also considered as one of the main factors leading to sub-optimality of the trajectory.

In this paper, these aspects are considered simultaneously. Uniform B-spline is utilized to represent trajectory and convert the trajectory generation problem into control point placement. A time-efficient non-uniform kinodynamic search method based on Euclidean signed distance field (ESDF) is proposed, ensuring the safety and reasonable time-allocation, and achieving optimality in time duration and control cost. Main contributions of this paper are summarized as follows:

\begin{itemize}

\item A uniform B-spline trajectory representation for quadrotors. A highly strict convex hull property is derived for a nonconservative dynamical feasibility check.
\item A time-efficient non-uniform kinodynamic search algorithm based on ESDF, which generates smooth, safe and reasonably time-allocated trajectories.
\item Implementation on simulator and hardware, for analyzing and verifying the performance. The source code of this work will be released on \url{https://github.com/tlb9551/BNUKsearch}.

\end{itemize}
\begin{figure}
\centering
\includegraphics[width=6cm]{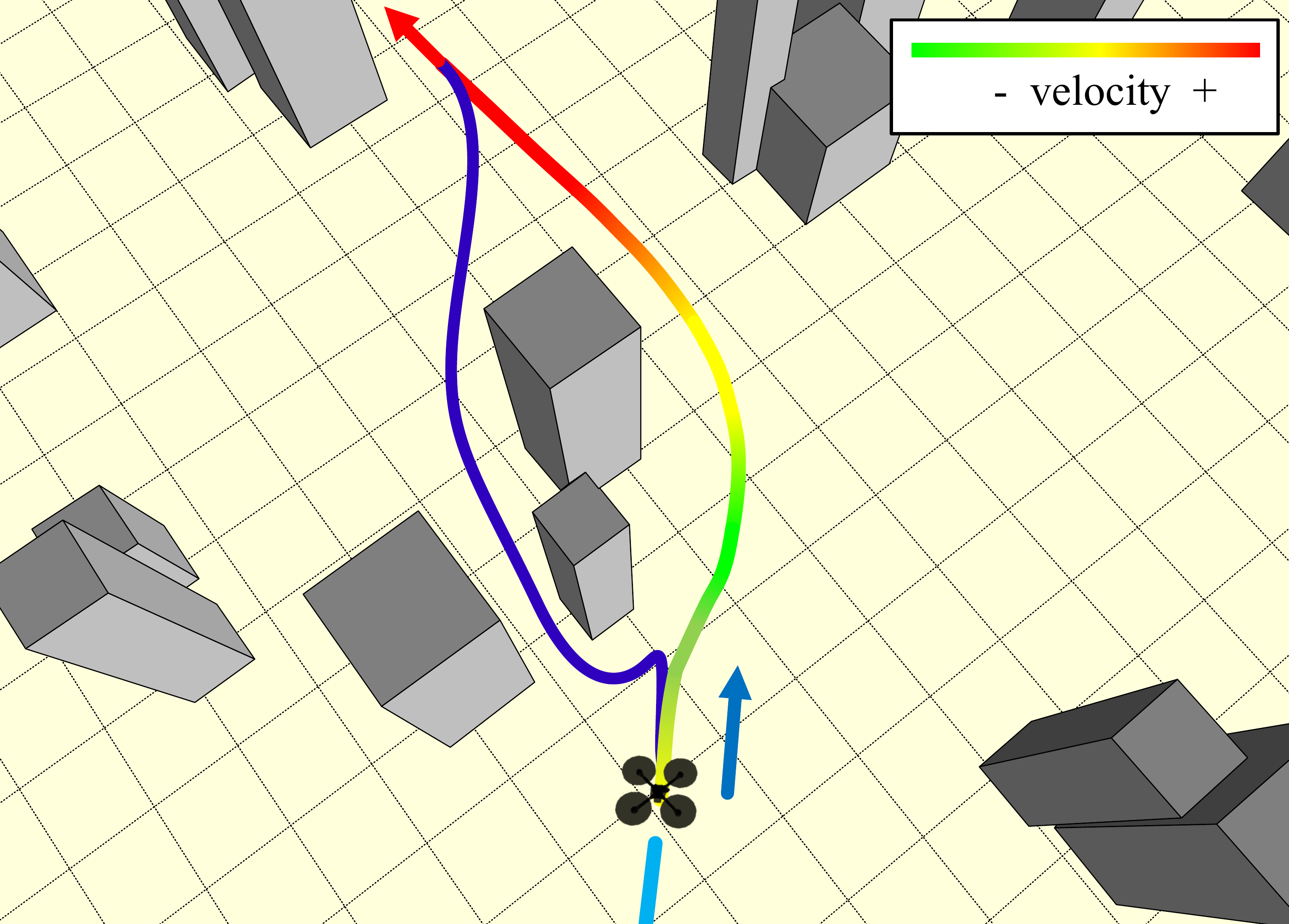}
\caption{Traditional method only considers the position and generates a trajectory (blue curve) from the shortest path, but facing the danger of dynamical infeasibility and insecurity. In contrast, the method proposed in this paper obtains a dynamically feasible and safe trajectory (colored curve). In addition, the trajectory is reasonably time-allocated according to the distance from the obstacle.}
\label{fig_sim}
\end{figure}


\section{ Related Works}
Trajectory generation for quadrotor can be transformed into the generation of the time-parameterized curve due to its differentially flat \cite{c_2_1}. When collision avoidance is considered in cluttered environments, curves such as B-splines \cite{c_2_2}, B\'{e}zier \cite{c_2_4} and piecewise polynomials \cite{c_2_5,wang_2} are used to represent shape-constrained trajectories.

Many existing planning methods take a two-step pipeline, i.e. a collision-free path is planned at first, then the smoothness and time-allocation of the relevant trajectory are optimized based on the shape of the path. At the front-end, sampling-based \cite{c_2_6,wang_1} and searching-based \cite{c_2_7,c_2_8} methods are used to plan a collision-free path. In the back-end, gradient-based methods \cite{c_2_9} and several other methods \cite{c_2_4,c_2_10} are employed to guarantee the smoothness and dynamical feasibility. However, these methods separate the trajectory shape and trajectory parameterization, are susceptible to some problems. For example, global optimal trajectory or even feasible trajectory may not be inside the homology class of path which is generated by the front-end methods without dynamic consideration.

\begin{figure}
\label{fig1}
\centering
\subfigure[]{\includegraphics[width=4.2cm]{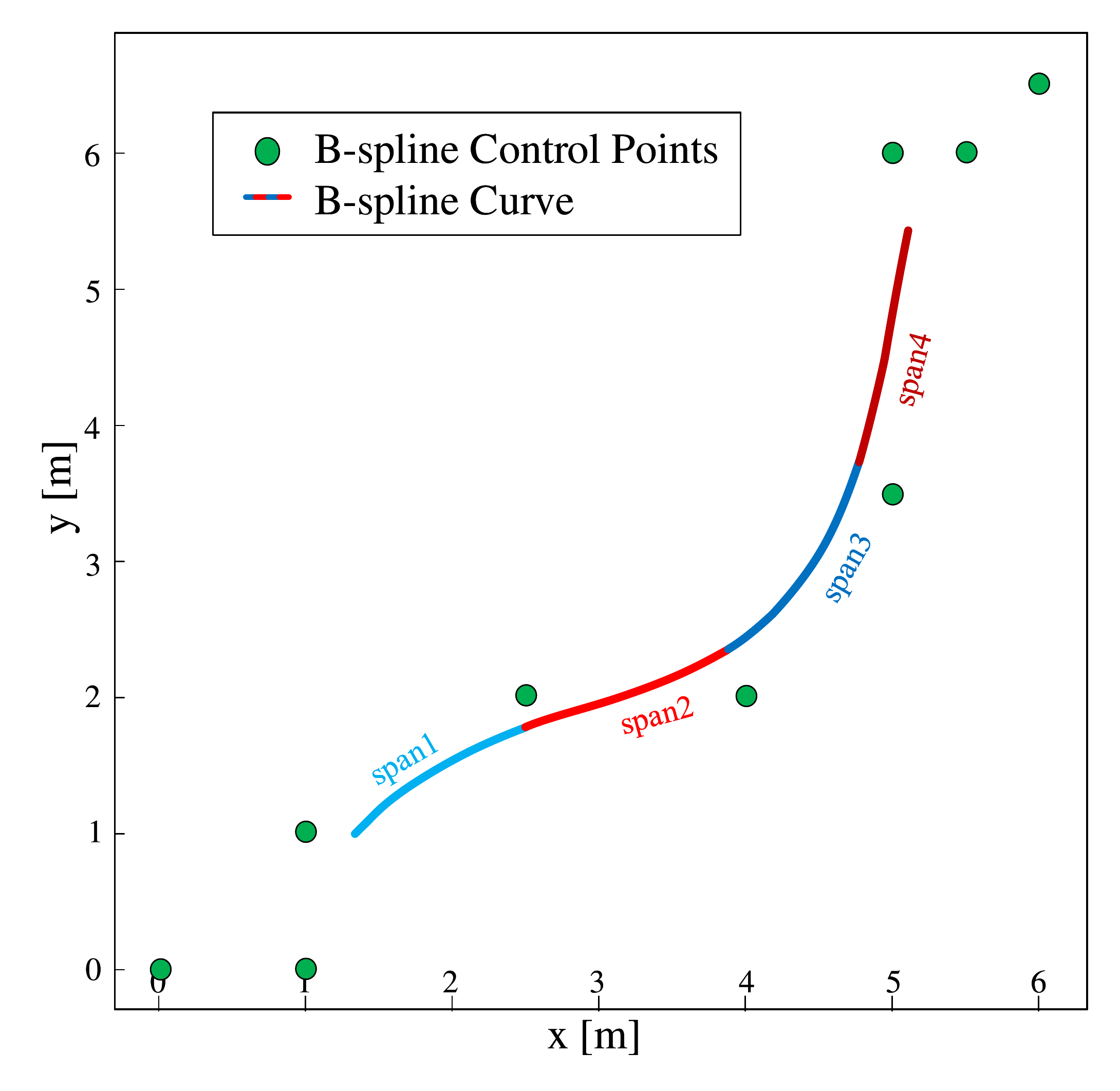}}
\subfigure[]{\includegraphics[width=4.2cm]{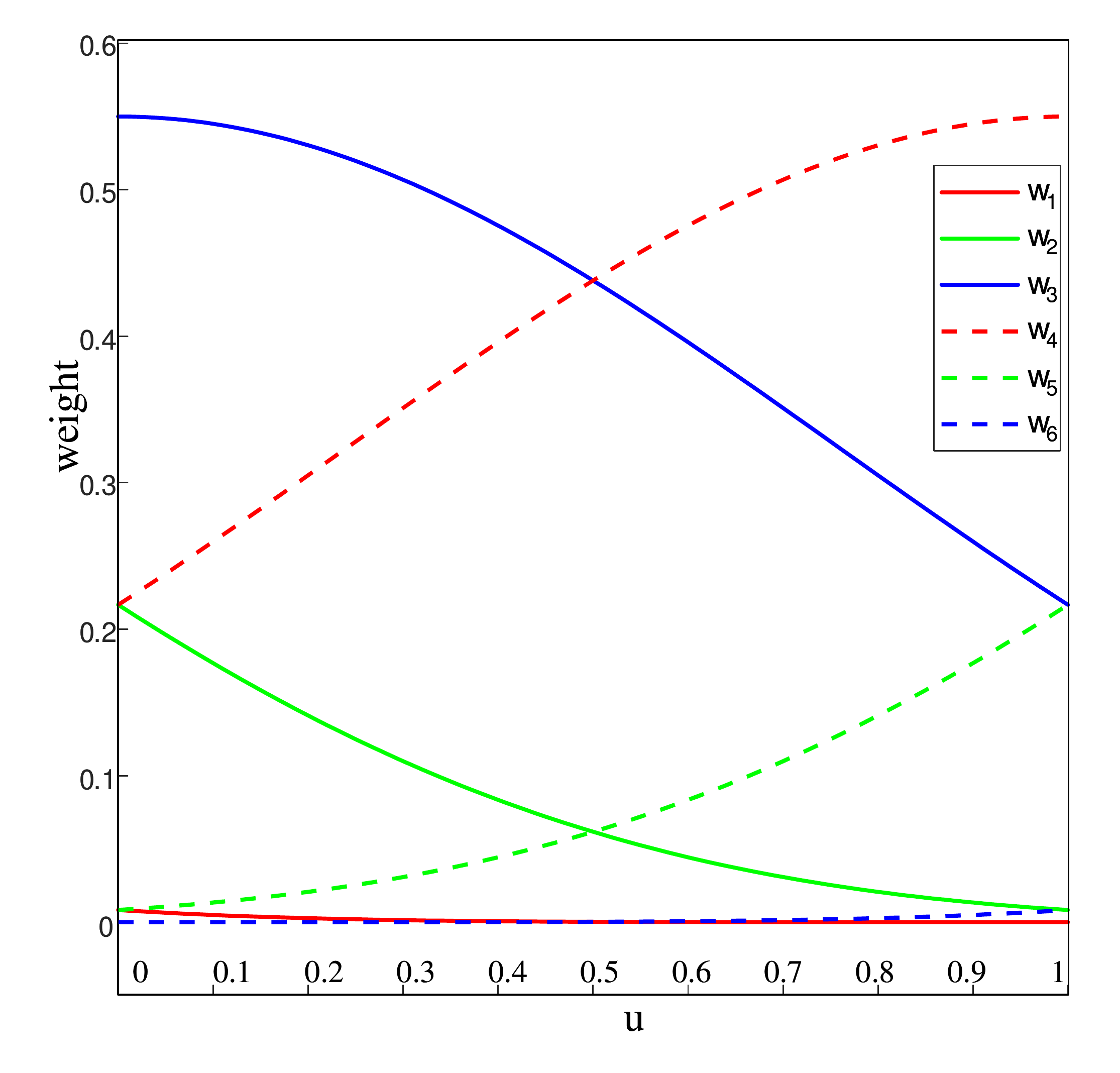}}
\subfigure[]{\includegraphics[width=4.2cm]{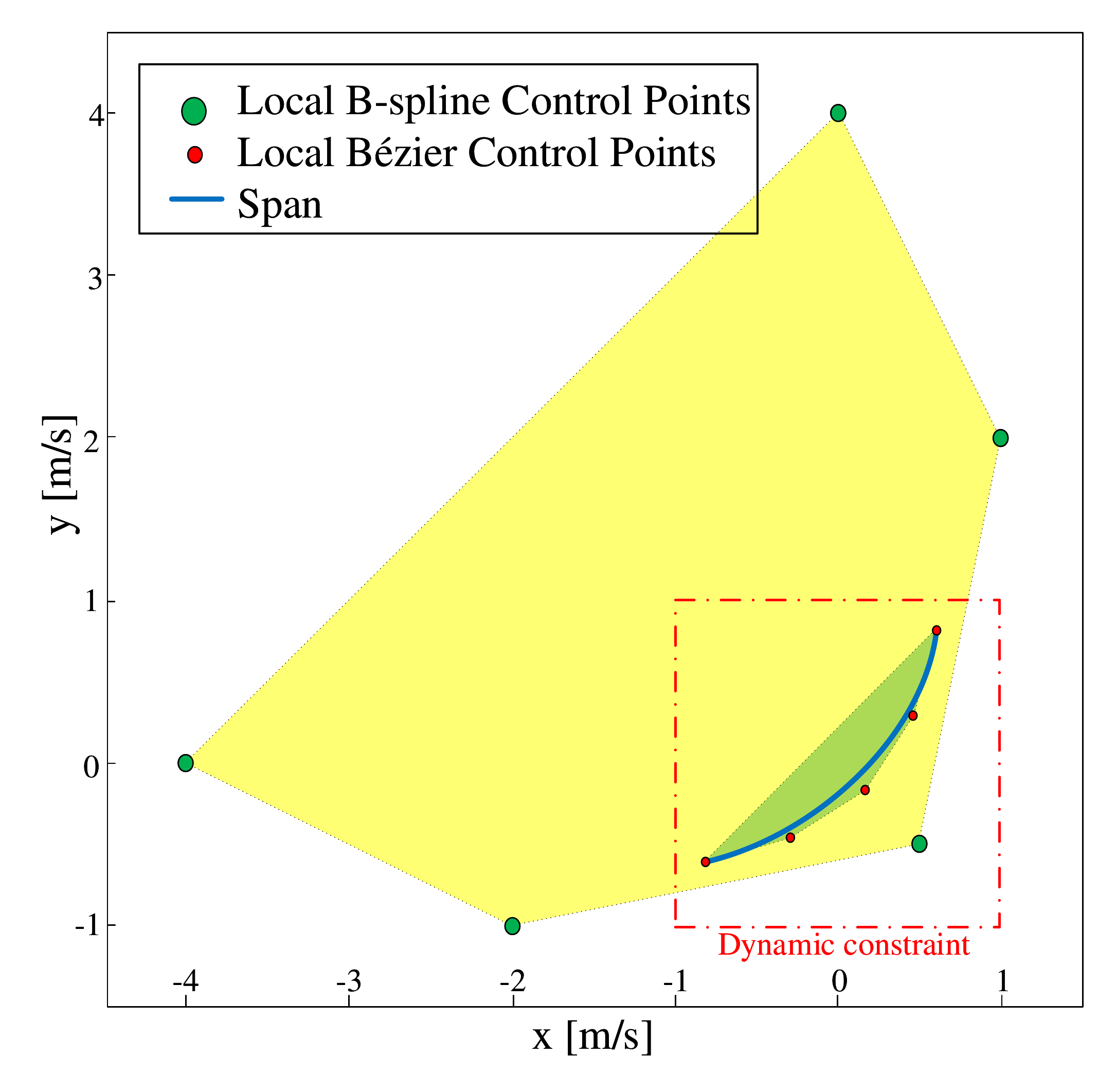}}
\subfigure[]{\includegraphics[width=4.2cm]{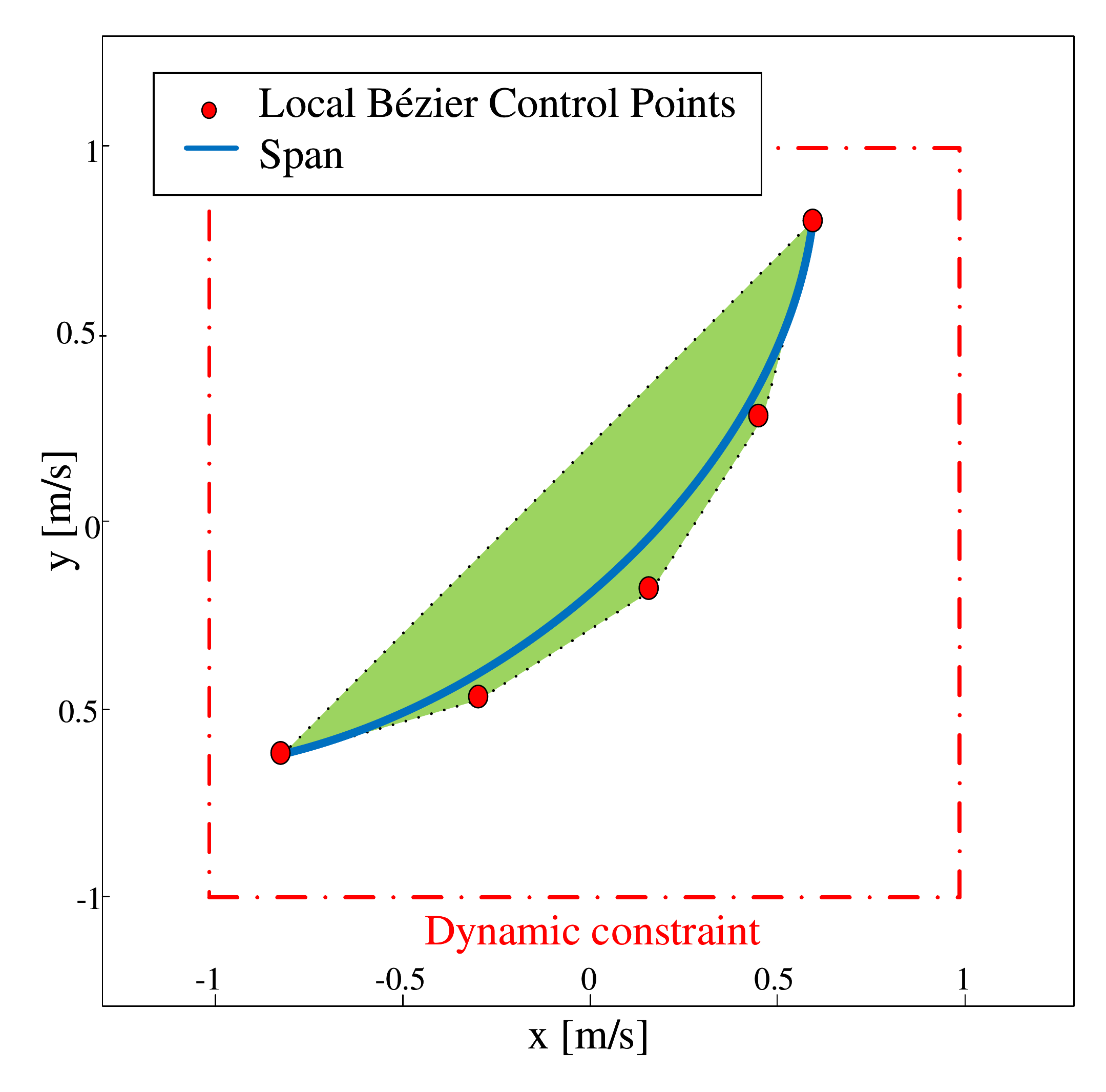}}
\caption{In (a), we show a quintic B-spline curve with nine control points, which has four spans, and each span is defined by its six consecutive local control points. (b) shows the weights of the six local control points that contribute to the span. In (c) and (d), we show the B-spline convex hull (yellow area) applied in existing work \cite{c_2_13} and the strict convex hull (green area) derived from the present paper. Due to the conservativeness of B-spline convex hull property (yellow area), a velocity curve may be misjudged as violating the dynamic constraint. However, the convex hull we derived (green area) closely wraps the curve, making the dynamical checking more accurate. }
\label{fig_2}
\end{figure}

Some methods consider both the trajectory shape and the dynamics in parallel. Usenko et al. \cite{c_2_12} used an optimization-based method that integrated the smoothness, dynamic constraints and obstacle-avoidance into the cost function of optimization but faced a low success fraction. Liu et al. \cite{liu2017search} proposed a primitive-based search method, the dynamical feasibility was guaranteed but met the problem of time-efficiency when high-order control inputs were required.

The time-allocation of trajectory is another significant factor that affects the quality of trajectories. Ding et al. \cite{c_2_13} refined the control points of B-spline while the total duration remains fixed. Fernbach et al. \cite{c_2_14} sampled the duration of segments until the related quadratic program is solved, but it may be not suitable to generate trajectories with many segments. Gao et al. \cite{c_2_4} generated time-indexed paths according to the density of obstacle, but the final trajectories do not strictly follow the heuristics after optimization. In this paper, we simultaneously deal with kinodynamic search and time-allocation of the trajectory, which shows noticeable benefits.

\section{B-spline based Trajectory Representation}\label{section:3}

Since the snap of the trajectory of quadrotor should be continuous \cite{c_2_1}, quintic uniform B-spline is selected to represent the trajectory of quadrotors. The B-spline is smooth and has some useful properties, which are suitable for our non-uniform kinodynamic search process. 

\subsection{Local Control Property and Kinodynamic Search} 
Search-based trajectory generation methods need to evaluate the cost of candidate trajectory increments frequently, and the local properties of B-spline curves are tailored to this. The value of a B-spline curve of degree $k-1$ can be evaluated by the following equation:
\begin{equation}
\bm c(t) = \sum\limits_{i = 0}^n {{\bm p_i}{B_{i,k}}(t)}
\end{equation}
where ${\bm p_i} \in {{\rm{\mathbb{R}}}^3}$ are the control points corresponding to ${t_i},i \in \{0,...,n\}$ and ${B_{i,k}}(t)$ are the blending functions which can be computed by the De Boor-cox recursive formula \cite{c_2_16}. 

Note that for uniform B-spline, the time interval between ${t_i}$ is fixed to be $\Delta t$, and the ${B_{i,k}}(t)$ is no-zero only in the interval of $t \in [{t_i},{t_{i + k - 1}})$, which means that a curve span in $t \in [{t_i},{t_{i + 1}})$ only depends on local $k$ control points $[{\bm p_i},...,{\bm p_{i + k - 1}}]$, it is called local control property. Therefore, a quintic B-spline curve with $n$ control points can be divided into $n-5$ spans (Fig. \ref{fig_2} (a)).

We define $s(t) = (t - {t_0})/\Delta t$ as a uniform time representation, $u(t) = s(t) - i \in [0,1]$ as a normalized parameter corresponding to the $i$-th span of the curve ${\bm c_i}(u(t))$, where $i$ is the integer part of $s(t)$. Therefore, the B-spline curve can be evaluated by the following matrix representation as mentioned in \cite{c_2_17}:
\begin{equation}
\label{eq2}
{\bm c_i}(u(t)) = {\bm b_6^ \top} {{\rm{M}}_6}{\bm P_i} 
\end{equation}
where ${\bm b_6^ \top}= [\begin{array}{*{20}{c}}1&u&{{u^2}}&{...}&{{u^5}}\end{array}]$ is the basis vector, ${\rm M_6}\in {\mathbb{R}^{6 \times 6}}$ is a constant basis matrix of quintic uniform B-spline and ${\bm P_i} = [\begin{array}{*{20}{c}}
{{\bm p_i}}&{{\bm p_{i + 1}}}&{...}&{{\bm p_{i + 5}}}
\end{array}]^ \top \in {\mathbb{R}^{6 \times 3}}$ is the vector of local control points of $i$-th span.

The derivatives of curve with respect to time can be also expressed as matrix representation. And the derivative curves are also B-spline curves with a lower degree, whose control points are a constant linear combination of higher degree B-spline’s control points. 

The integral over squared time derivatives of the $i$-th span, which expresses the control cost of quadrotor trajectory \cite{c_2_1}, can be also locally computed in closed form during the search process:

\begin{equation}
\begin{split}
\bm E_{_i}^c &= \sum\limits_{l = 1}^4 {\int_{{t_i}}^{{t_{i + 1}}} {{w_l}{{\left({\frac{{d{{\bm{c}}_i}(u(t))}}{{{d^l}t}}} \right)}^2}dt} }\\
&= \sum\limits_{l = 1}^4 {{w_l}{\bm P_i^\top} \mathrm M_6^\top{\mathrm Q_l}{\mathrm M_6}{\bm P_i}} 
\end{split}
\end{equation}
where ${Q_l} = \frac{1}{{{{\left({\Delta t} \right)}^{2l - 1}}}}\int_0^1 {\left({\frac{{d\bm b}}{{{d^l}u}}} \right){{\left({\frac{{d\bm b}}{{{d^l}u}}} \right)}^ \top }du}$ is a constant matrix and can be computed in advance. The local control property and the matrix representation of $\bm E_{_i}^c$ enable incremental calculation of the cost during the search process.

One understanding of Eq. \ref{eq2} is that a span of quintic B-spline can be regarded as weighted combinations of six local control points, and the weight is the time-varying basis, $\bm b^ \top_6(u){{\rm{M}}_6}$. Fig. \ref{fig_2} (b) shows the weight of each control point that contributes to a span when the uniform time parameter $u \in [0,1]$. The weights of ${\bm p_3}$ and ${\bm p_4}$ are much higher than other control points, so the span will be close to the connection line between ${\bm p_3}$ and ${\bm p_4}$ as shown in Fig. \ref{fig_2} (a). As a result, we can easily confirm that the trajectory mostly has no collision with obstacles as long as the connection line of successive control points does not pass through obstacles. Note that it is not a sufficient condition, and we will introduce the convex hull property to build a sufficient condition in the next subsection.

\subsection{Strict Convex Hull Property and Feasibility Checking}

The dynamical feasibility of trajectory guarantees that trajectory can be well tracked by the motion controller while some additional tasks are performed well \cite{wang_3}. As an important property of B-spline, the convex hull property can be used to conservatively evaluate the spans. Existing work is faced with limitations due to its conservativeness, but the novel strict convex hull property we derived here can be used to evaluate the spans of B-spline in a nonconservative way, so as to assess more accuracy in feasibility checking.
 
The convex hull property of B-spline means that a span of B-spline ${\bm c_i}(u)$ lies within the convex hull of its local control points ${\bm p_{i + j}} (j = 0,...,k - 1)$, where $k - 1$ is the degree of B-spline (see Fig. \ref{fig_2} (c)). Since the derivatives of a B-spline curve are also B-spline curves with a lower degree \cite{c_2_16}, they share the same property: local control property and convex hull property. Thus the dynamical feasibility and collision-free constraints can both be expressed as bounds on the placement of control points.

In the previous work \cite{c_2_13}, the convex hull property of B-spline is directly used to check the dynamical feasibility, a span is judged to be dynamically infeasible unless all the local control points of its derivative curve are within the border of the dynamically feasible domain, i.e. $\left\| {\bm p_{_{i + j}}^\prime} \right\| < {v_{\max }}$ and $\left\| {\bm p_{_{i + j}}^{\prime\prime}} \right\| < {a_{\max }}$. However, since the convex hull of uniform B-spline control points does not tightly wrap the span, this method has some limitations of conservativeness as shown in Fig. \ref{fig_2} (c).

To solve this problem, we try to find a linear transformation of local control points ${\bm Q_i}={\rm{L}}{\bm P_i}$, so that the convex hull formed by ${\bm Q_i}$ more tightly wrap the span. We noticed that the B\'{e}zier curve also has convex hull property and the B\'{e}zier curve is guaranteed to pass through the first and the last control points, so we transform the local span of B-spline into a B\'{e}zier curve representation to achieve tighter convex hull(Fig. \ref{fig_2} (d)).
Given:
\begin{equation}
{\bm c_i}(u) = {\bm b_k^ \top} {{\rm{M}}_k}{\bm P_i} = {\bm b_k^ \top }{{\rm{B}}_k}{\bm Q_i}
\end{equation}
Thus:
\begin{equation}
\label{eq8}
{\bm Q_i} = {{\rm B}_k}^{ - 1}  {{\rm M}_k}{\bm P_i}
\end{equation}
where ${\rm B_k}$ is the constant basis matrix of B\'{e}zier curve, and ${\bm Q_i}$ are the local control points of B\'{e}zier curve. Thus we can use the strict convex hull property for dynamically feasible checking as follows.

\emph{\underline{Dynamical feasibility checking}}: firstly, the control points of the derivatives curve of trajectory are evaluated. Then we computed the corresponding B\'{e}zier control points with Eq. \ref{eq8}. Finally, if all the B\'{e}zier control points are in the dynamically feasible domain, i.e. $\left\| {\bm q_{_{i + j}}^\prime} \right\| < {v_{\max }}$ and $\left\| {\bm q_{_{i + j}}^{\prime\prime}} \right\| < {a_{\max }}$, the span of B-spline trajectory is judged to be dynamically feasible (Fig. \ref{fig_2} (d)).

\begin{figure}
\centering
\includegraphics[width=6.4cm]{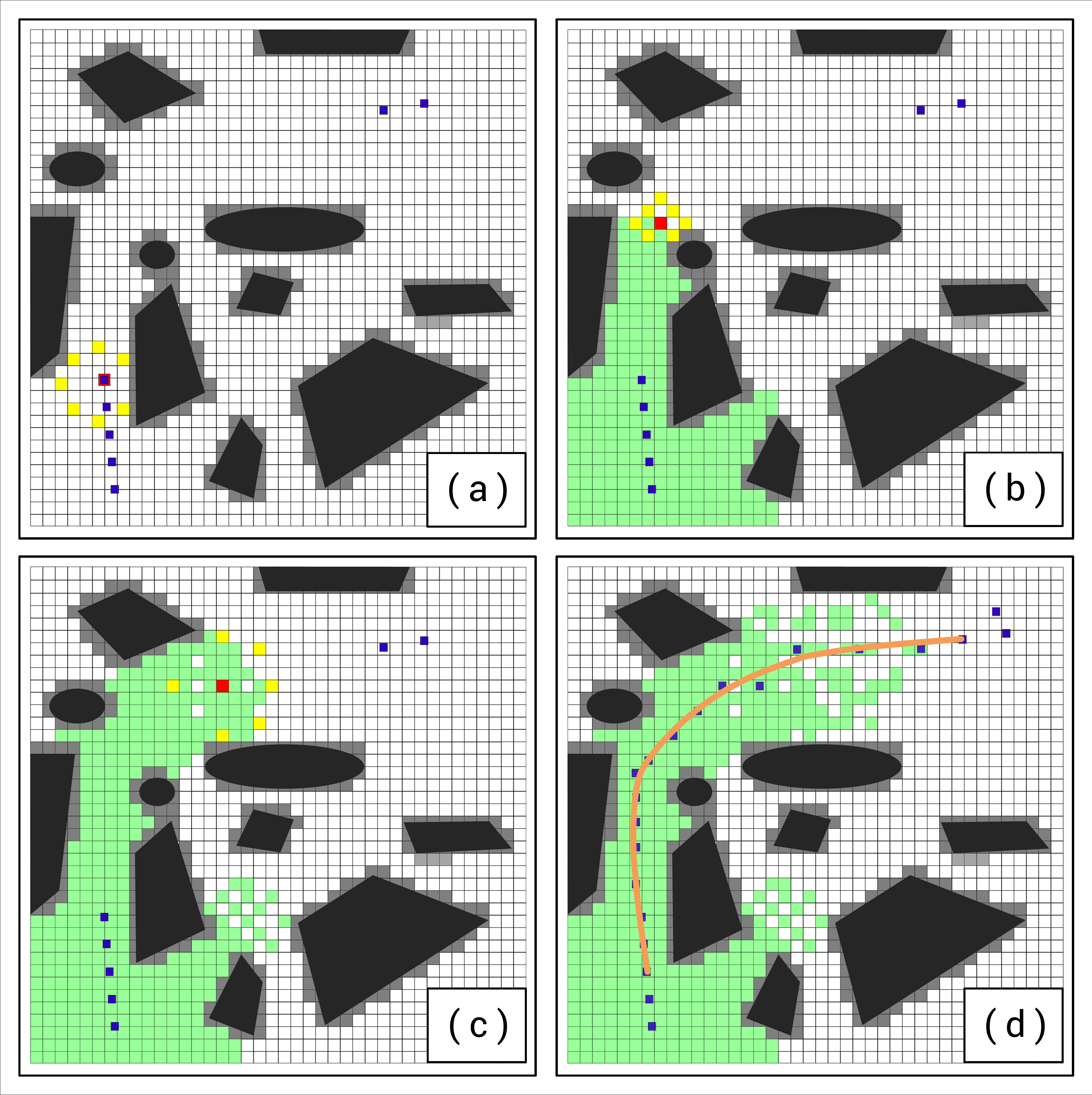}
\caption{Our search process starts from the starting nodes (blue points). The node with the lowest cost is set as the current node (red point) and is expanded to find its neighboring nodes (yellow points) according to the distance from obstacle as shown in (b) and (c), During the search process, dynamically infeasible nodes are excluded. As soon as reaching the goal area, the search process is terminated and the control points (blue points) of trajectory are retrieved, and the B-spline trajectory is calculated with the control points as in (d).}
\label{fig:3}
\end{figure}

\subsection{Non-static Initial and Goal State}
In order to keep the trajectory continuous up to a higher derivative of the quadrotor when the replanning happens in flying, the non-static initial state should be allowed. 
Given the non-static state of the quadrotor, position and its derivative, $\left\{ {\bm p,{\bm p^\prime},{\bm p^{\prime\prime}},{\bm p^{\prime\prime\prime}}} \right\}$ as the beginning of trajectory, the first five control points can be obtained by solving the following equation set:
\begin{equation}
{\left. {\frac{{d{{\bm{c}}_i}(u)}}{{d{t^l}}}} \right|_{u = 0}} = {\left. {{\bm{b}^ \top _{6 - l}}} \right|_{u = 0}}{{\rm{M}}_{6 - l}}{\rm {S}_{l}}{\bm P_i} = {\bm p^{(l)}},l = 0,1,2,3
\end{equation}
and the 6-th control point is arbitrary.

The goal control points can also be calculated by a similar approach. In our implementation, the goal state is set as a waypoint with specified velocity. Therefore, only two equations related to position and velocity, confirm the last two control points of the whole trajectory.

\section{Non-uniform Kinodynamic Search}

\begin{algorithm}
\caption{BNUK search($s_{start}$, $s_{end}$, ${\cal G})$}
\label{alg:bnuk}
$p_{start} \Leftarrow$ ComputeStartNode($s_{start}$)\;
$p_{end} \Leftarrow$ ComputeEndNode($s_{end}$)\;
${\cal O}={\cal I}=\emptyset$\;
Add(${\cal O}$,$p_{start}$,Cost($p_{start}$))\; 
\While {\rm{Size(${\cal O}\neq 0$)}}{
$p_{cur} \Leftarrow $ PopMin(${\cal O}$), Add(${\cal I}$,$p_{cur}$)\;
\If{$p_{cur}=p_{end}$}{
 $CPs \Leftarrow$ Retrieve($p_{cur}$,$all$)\;
\Return EvaluteTraj($CPs$)\;
}
\For{$p_{nbr} \in $ \rm{\underline{NodeExpansion}}($p_{cur}$) $\cap$ $p_{nbr} \notin {\cal I}$}{
 $LocalCPs \Leftarrow$ Retrieve($p_{nbr}$,$6$)\;
\If{\rm{\underline{CheckDynamic}}($LocalCPs$)$ = true $} {
 $c_{ten}=$ Cost($p_{cur}$) $+$Cost(${LocalCPs}$) $+\Delta t$\;

\If{$p_{cur} \in {\cal O}$} {
\If{$c_{ten} < $ \rm{Cost($p_{nbr}$)}}{
 Update($p_{nbr}$,$c_{ten}$)\;
}
}
\Else{
 Add(${\cal O}$,$p_{nbr}$,$c_{ten} +$\underline{HeuristicCost}($p_{nbr}$))\;
 }

}
}
}
\end{algorithm}

As discussed in the previous section, the whole B-spline trajectory can be expressed as a set of B-spline spans. Successive spans share some same control points, i.e. the last $k - 1$ control points of span ${\bm c_i}$ are the first $k - 1$ control points of ${\bm c_{i + 1}}$. The increase in control points leads to an increase in the spans and total trajectory duration. Therefore, the trajectory planning is transformed as a problem to find a set of B-spline control points, which generates a collision-free, dynamically feasible and cost-optimized B-spline trajectory between start and goal state. Such a problem can be treated as a graph optimization problem and solved by our search-based algorithm. For convenience, we call it BNUK (B-spline based Non-uniform Kinodynamic) search (Alg. \ref{alg:bnuk}).

\subsection{Algorithm Overview}

The graph structure ${\cal G}:=({\cal V}{,}{\cal E})$ is chosen as follows. Vertices ${\cal V}$ is the cell center of voxels in discretized 3-D space, indicating control points, and edges ${\cal E}$ is their connections. Each vertex is connected to its neighbors in 26 directions like the usual A* search algorithm in 3-D space but obeys the rule of choosing specific step length between neighbors, which will be exhaustively discussed in the next subsection. 

Thus the kinodynamic search problem is to find an ordered set of control points ${\cal S}: = \left\{ {{\bm p_1},{\bm p_2},...,{\bm p_n}} \right\}$ from the graph ${\cal G}$, which satisfy the dynamical feasibility, and minimizes the cost of trajectory as given below,
\begin{equation}
\mathop {\rm {min}}\limits_{{\cal S}\subset{\cal V}} J({\cal S}) = \sum\limits_{i = 0}^{n - k + 1} \lambda {\bm E_i^c} + (n - k + 1)\Delta t
\end{equation}
which is a weight of control cost and trajectory duration. The cost can be evaluated incrementally due to the local control property.

\subsection{Non-uniform Node Expansion}

Node expansion refers to finding neighboring nodes of the current node and is implemented in function $NodeExpansion({p_{cur}})$ in Alg. \ref{alg:bnuk}.

Different from traditional path finding as A* search, the output of BNUK search, i.e. a set of control points ${\cal S}$ is not required to be connected in 3-D discrete space. Therefore, we propose a novel non-uniform node expanding strategy. The step length of the node expansion, i.e. the spacing between two control points, is dynamically adjusted during the search process.

As discussed in Sec. \ref{section:3}, the spacing length between control points will affect the derivatives of the trajectory due to the fixed time interval $\Delta t$. We argue that it is reasonable to reduce the velocity of quadrotors in the vicinity of the obstacles for security consideration, because the actual trajectory is always deviated from the planned trajectory due to the existence of control errors. Similarly, the velocity should be increased to take full advantage of quadrotor’s dynamics when away from obstacles. Therefore the spacing length between control points needs to be set in reference to some quantity describing the distance from the obstacle.

The Euclidean signed distance field (ESDF) is employed as the reference. The cells of ESDF store the Euclidean distance to the nearest obstacle. We use the following function $f(d)$ to mapping distance $d$ to step length of node expansion:
\begin{equation}
f(d) = \left\{ \begin{array}{l}
0,\\
d - \tau,\\
{v_{\max }}\cdot \Delta t,
\end{array} \right.\begin{array}{*{20}{c}}
\begin{array}{r}
d \le \tau \\
\tau < d \le \tau+{v_{\max }}\cdot \Delta t\\
d \ge \tau + {v_{\max }} \cdot \Delta t
\end{array}
\end{array}{\rm{ }}
\end{equation}
where $\tau$ is a parameter related to the radius of the quadrotor. This function is limited under ${v_{\max }}\cdot \Delta t$, so it does not make the spacing between control points too large to cause the trajectory exceeding the dynamical velocity limit. And step length is proportional to the distance and is equal to 0 when the distance is under $\tau$, meaning that the node is always expanding in the collision-free area. This avoids time-consuming obstacle inflation and also improves safety.

The non-uniform node expansion benefits the search process in three aspects. Firstly, safety is improved because the step length is always shorter than the distance to the nearest obstacle, no control point will be set in non-free or dangerous area. Secondly, since the trajectory duration is a part of cost function, equal to ${(n - k + 1)\Delta t}$, expanding to a node with large step length is less costly than a node with small step length. Thus the trajectory will be naturally away from obstacles to achieve less trajectory duration. Thirdly, trajectory is reasonably time-allocated, slowly in obstacle-dense area and fast in obstacle-sparse areas. 

\begin{figure}
\centering
\subfigure[A* search]{\includegraphics[width=3.2cm]{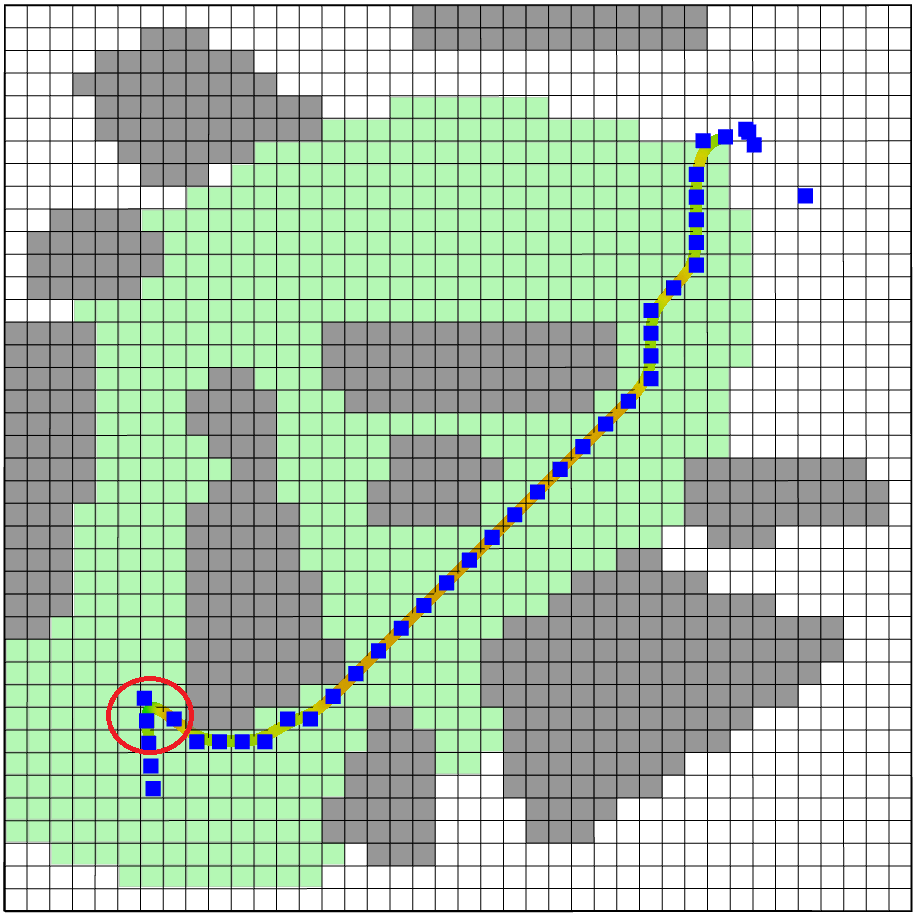}}
\subfigure[RBK search]{\includegraphics[width=3.2cm]{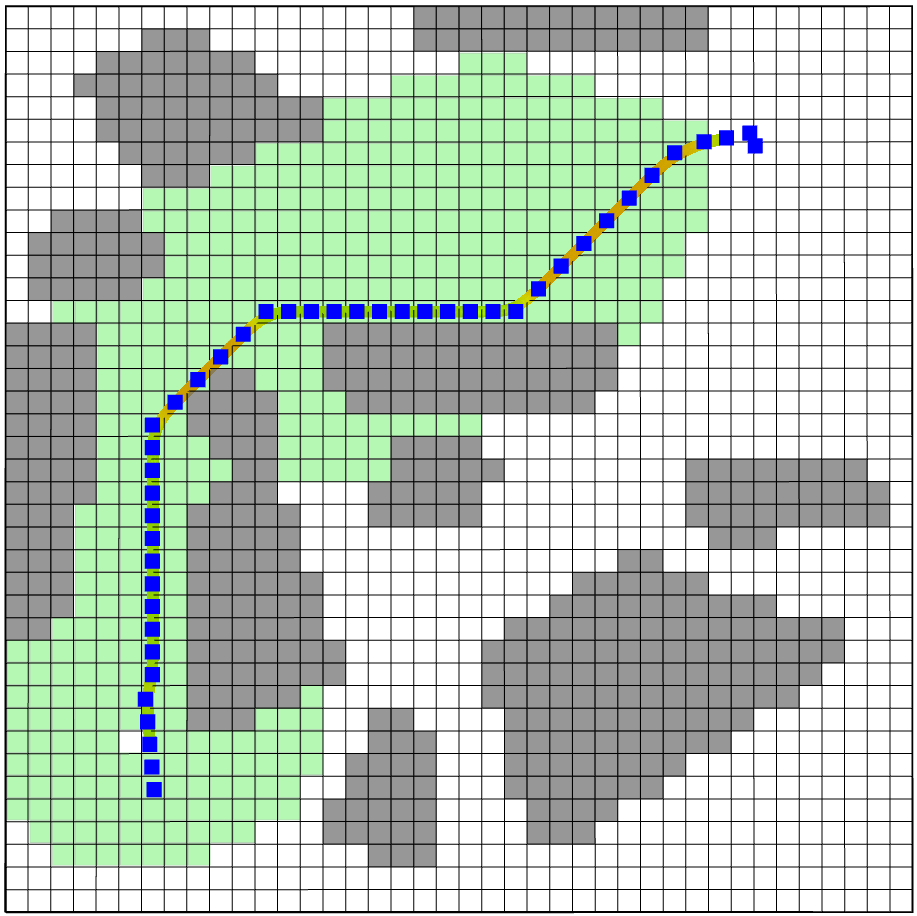}}
\subfigure[our BNUK search]{\includegraphics[width=3.2cm]{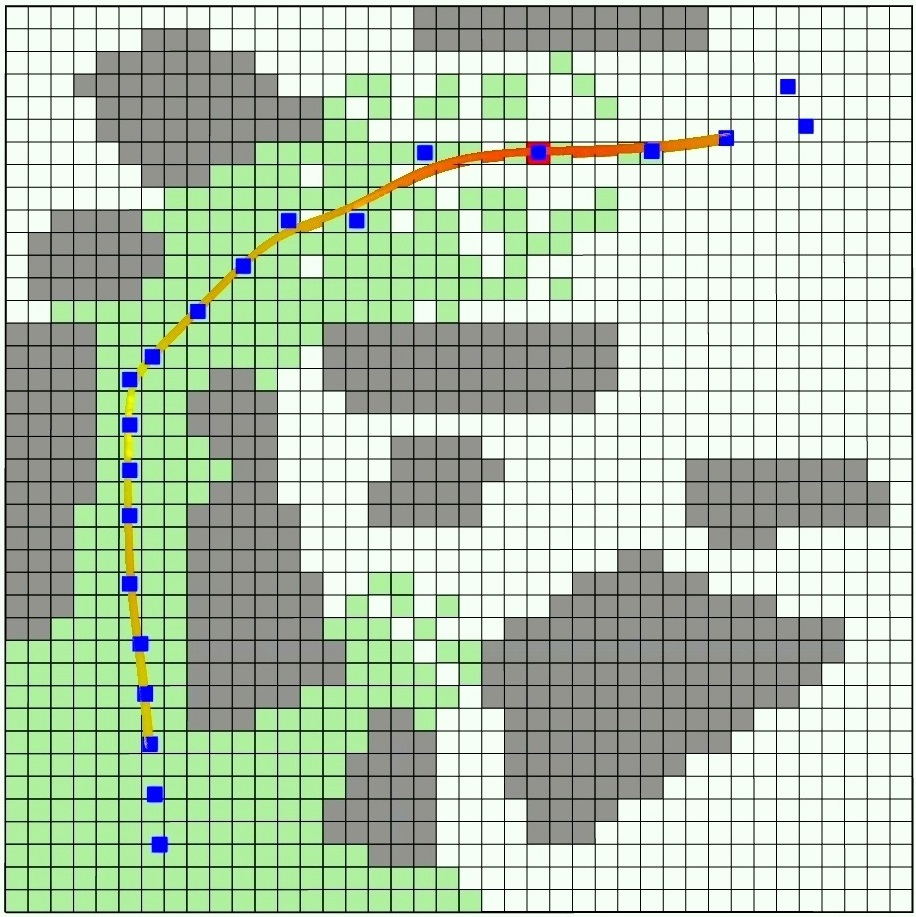}}
\subfigure[Replanning in large map]{\includegraphics[width=3.2cm]{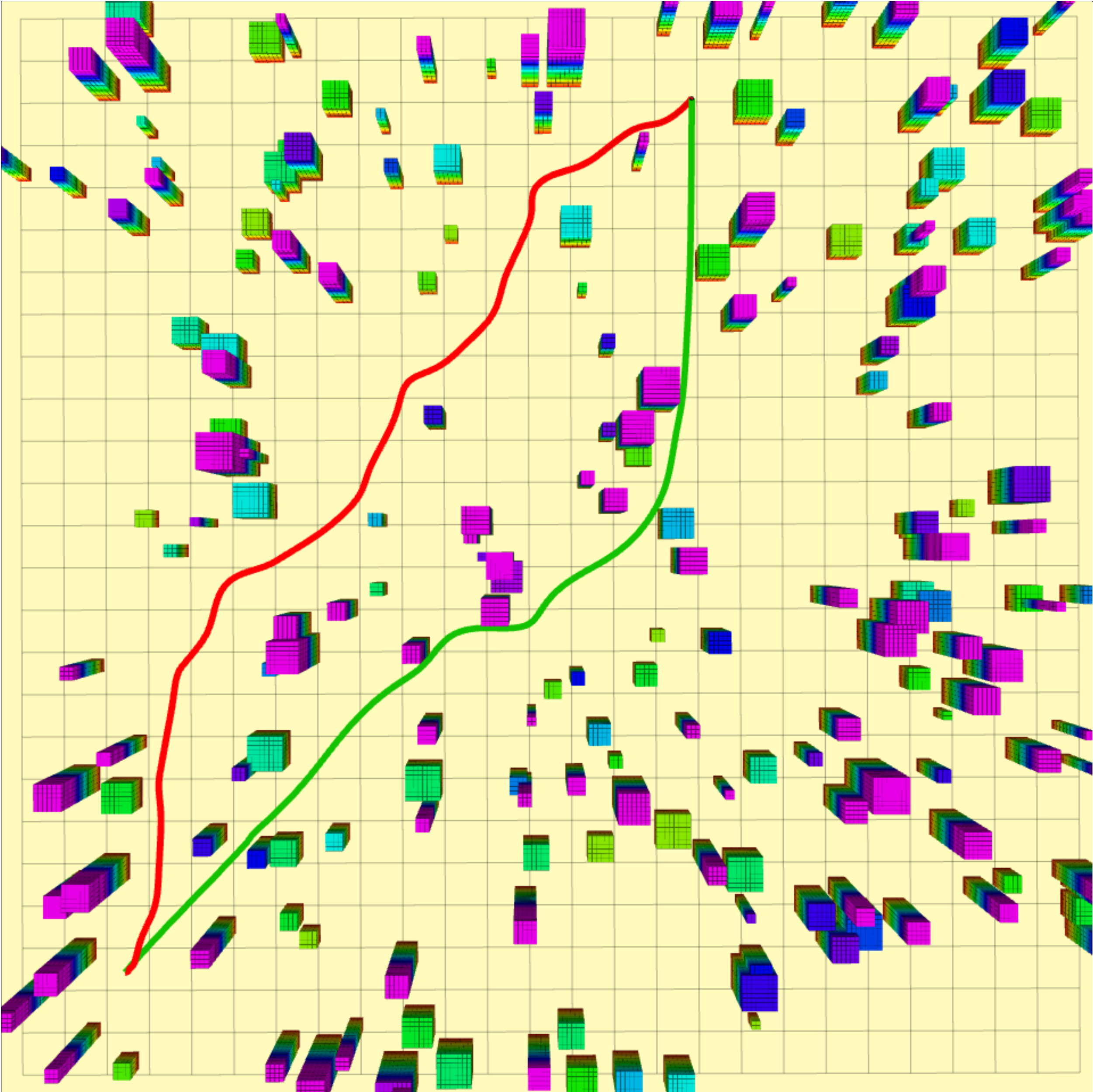}}
\caption{(a), (b) and (c) are the illustrations of comparison between A* search, RBK search, and our BNUK search. As in (a), A* search generates a trajectory with the shortest length but takes the risk of being dynamically infeasible (in red circle) due to the non-zero initial velocity. In (b), the RBK search avoids the dynamical infeasibility by dynamical checking during searching, but the trajectory is too close to obstacles and needs time-allocation. In (c), our BNUK search obtains a safe and dynamically feasible trajectory, and the trajectory is reasonably time-allocated according to the obstacle-density. (d) shows our trajectory (red curve) is safer than the trajectory (green curve) of Gao's method \cite{c_2_4} as our trajectory keeps a safe distance from obstacles during the replanning simulation.}
\label{fig_sim}
\end{figure}

\subsection{Dynamical Feasibility Checking and Heuristic Function}

The dynamical checking and cost evaluating both need the value of local control points, thus we utilize the function $Retrieve()$ to get the control points of the local span. Then dynamical checking and cost evaluating are done by the method discussed in Sec.\ref{section:3}. As for the collision-free checking, it is time-consuming and unnecessary so we leave it out of the search process.

The heuristic function is useful to speed up the search process due to the reduction of expanded node number, it is an estimation of cost from the current node to the goal:
\begin{equation}
HeuristicCost({p_{nbr}}) = DiagDis({p_{nbr}},{p_{goal}})/{v_{\max }}
\end{equation}
where $DiagDis()$ denotes the diagonal distance. This heuristic function is admissible since it underestimates the optimal cost from $p_{nbr}$ to $p_{end}$, and is efficient in speeding up the search process.

\section{Experiments}

\begin{figure}
\centering
\subfigure[Gao's method]{\includegraphics[width=8cm]{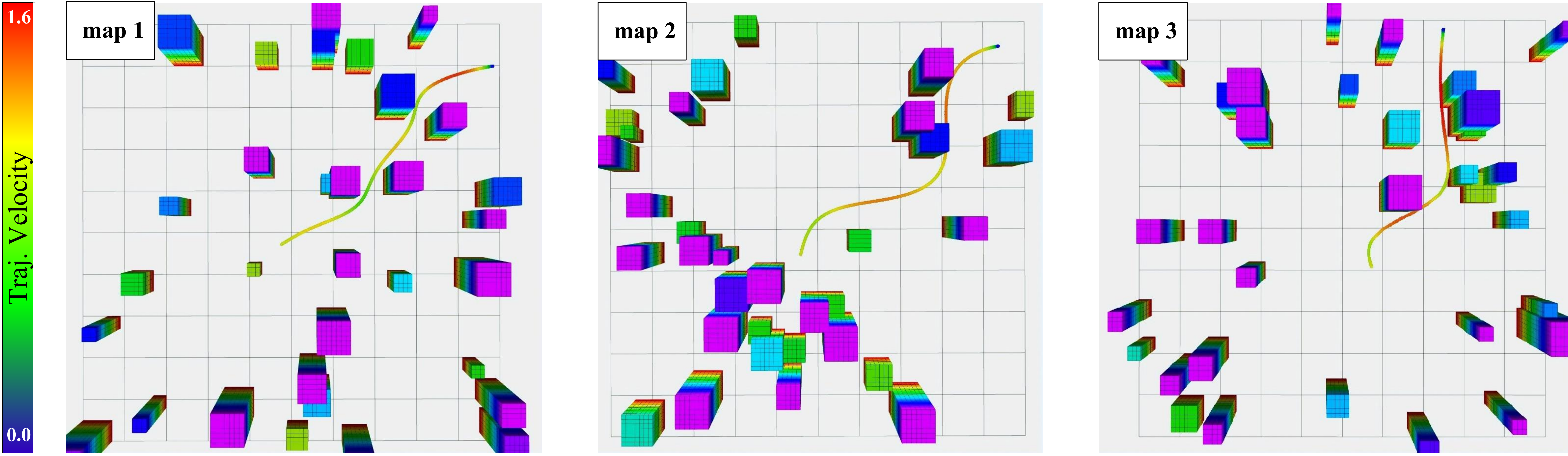}}
\subfigure[Our method]{\includegraphics[width=8cm]{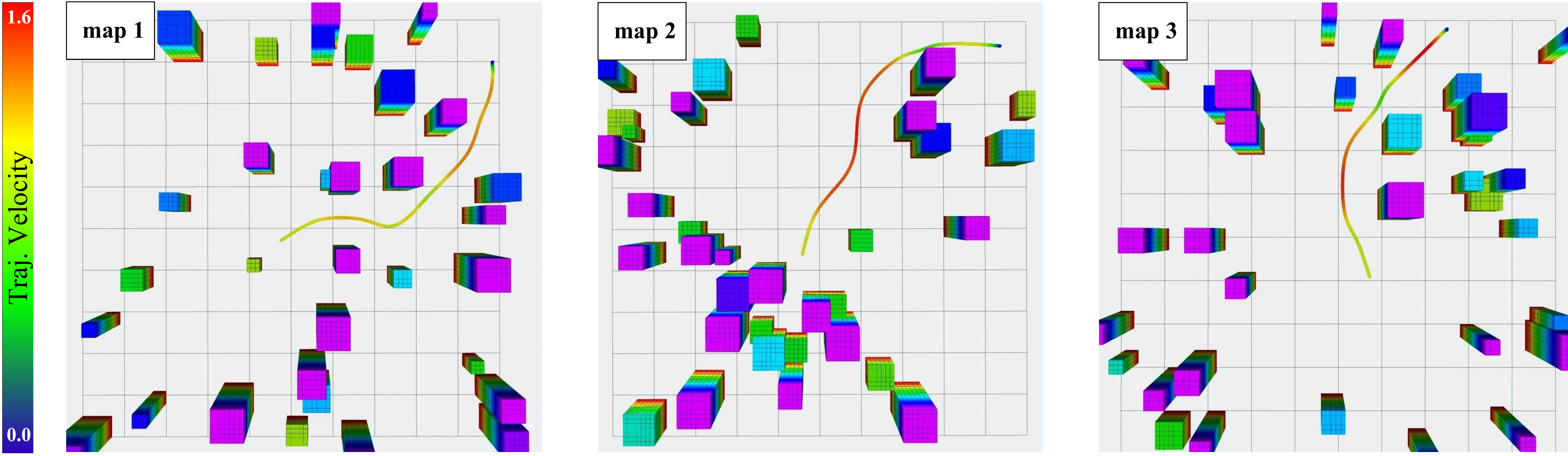}}
\caption{Trajectory generation in random maps. Both methods generate trajectories with reasonable time-allocation, but our method is safer, smoother and more time-efficient.}
\label{fig:6}
\end{figure}
\subsection{Comparison with State-of-the-Art Method}
In this section, we compared our BNUK search with a state-of-the-art method proposed by Gao et al. \cite{c_2_4} in two ways, trajectory generation on random maps and navigation simulation on large maps. 
\begin{table}
\caption{Comparison of Trajectory Generation \protect\\ on 50 Random Maps}
\tiny
\label{table_1}
\begin{center}
\begin{tabular}{|c||c|c|c|c|c|c|c|c|}
\hline
 &Traj.&Traj.&Comp.&Avg.&Avg.&Min.&Avg. \\
 &Length&Time&Time&Vel.& Acc.&Dis. &Dis. \\
 &(m)&(s)&(ms)&(m/s)&(m/\rm{$s^2$})&(m)&(m)\\
\hline
Our method&12.95&11.69&\textbf{16.9}&1.11&0.46&\textbf{0.97}&\textbf{1.82}\\
\hline
Gao's method \cite{c_2_4}&12.39 & 11.94&68.56&1.04&0.29&0.54&1.63\\
\hline
\end{tabular}
\end{center}
\end{table}
Gao's method uses the fast marching in the velocity field transformed from ESDF to find a time-indexed path, the trajectories represented by piecewise B\'{e}zier curves are generated by solving a convex quadratic program. We compared our method with Gao's method because both methods use the obstacle density information in EDSF to generate trajectories with reasonable time-allocation, but in different ways and trajectory representation, resulting in different performance in time-efficiency and trajectory quality.

We generate 50 random maps and plan trajectories from the center to side and corner with two methods, simulating local replanning that often occurs in autonomous navigation. The map size is $20m \times 20m \times 4m$ and resolution is $0.2m$. The maximum velocity and acceleration are set as $1.6m/s$ and $1.6m/{s^2}$. As the result shown in Tab. \ref{table_1}, both methods generate dynamically feasible trajectories and behave similarly in terms of trajectory length and duration. The mean and minimum distance from obstacles of our methods are larger than Gao's method, meaning that the trajectory generated by our methods is safer. In addition, our method is much more time-efficient, and the average calculation time is less than a quarter of theirs.

We also simulate the navigation in several large random forest maps with two methods. RHC (Receding Horizon Control) \cite{c_2_18} is used as the same strategy. Replanning happens when the quadrotor arrives at the border of prior sensing horizon or when a collision is detected. The map size is $50m \times 50m \times5m$ and resolution is $0.2m$, the local sensing map size is $20m \times 20m \times 5m$. We set the same maximum velocity and acceleration as $1.6m/s$ and $1.6m/{s^2}$, same start and goal are set so that the trajectory length is longer than $50m$. As shown in Fig. \ref{fig_sim} (d), our trajectory (red curve) is safer than Gao's (green curve). Tab. \ref{table_2} shows the average data of two methods, the trajectory length is similar but our method reaches a lower duration. Our method is much more time-efficient. The replanning times of Gao's method are higher than ours because it encountered failures during the trajectory optimization process.
\begin{table}
\caption{Comparison of Replanning on Large Random Maps}

\label{table_2}
\begin{center}
\begin{tabular}{|c||c|c|c|c|}
\hline
 &Traj.&Traj.&Comp.&Replan\\
 &Length&Time&Time&Times\\
 &(m)&(s)&(ms)& \\
\hline
Our method&55.33&\textbf{46.95}&\textbf{44.6}&\textbf{8.6}\\
\hline
Gao's method \cite{c_2_4}&\textbf{51.75}&56.94&194.27&26.3\\
\hline
\end{tabular}
\end{center}
\end{table}
\begin{figure}
\centering
\includegraphics[width=7cm]{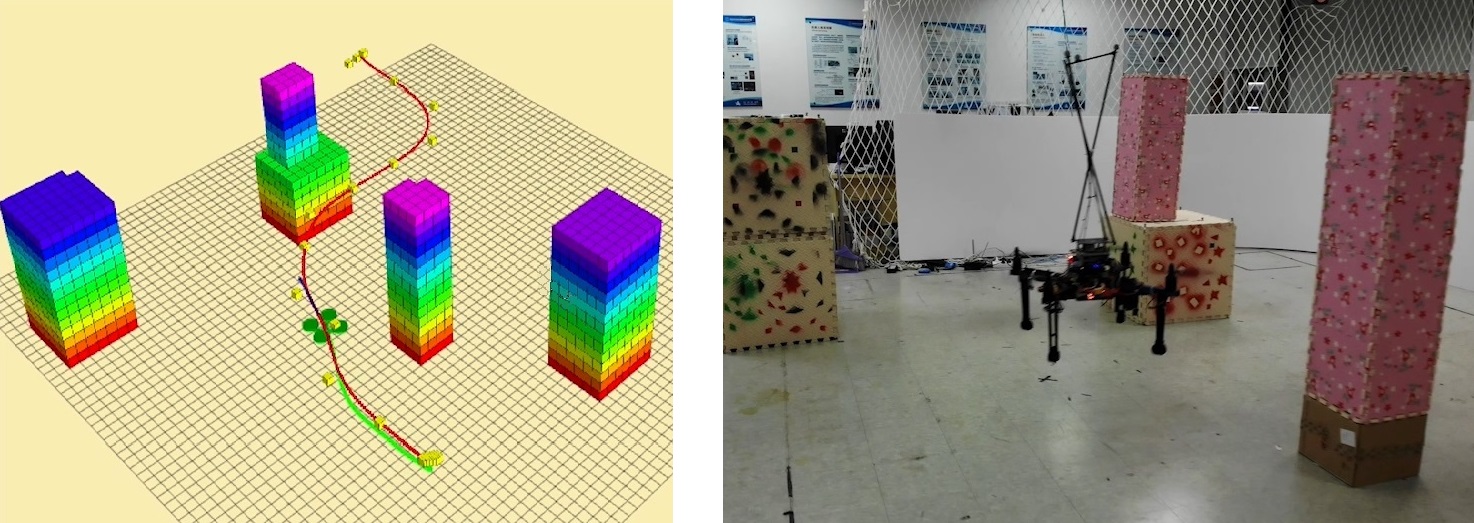}
\caption{Indoor experiment. The quadrotor generates a safe and smooth trajectory to avoid the obstacles while flying. The red curve indicates the trajectory planner by our method, the green curve is the ground truth.}
\label{fig:flight}
\end{figure}

\subsection{Flight Experiment}

Flight experiment is done to demonstrate the collision avoidance and real-time calculation of our method. We use a quadrotor based on the Pixhawk flight controller, and an UP2 embedded board is utilized as the onboard computer. A motion capture system is used to locate the quadrotor and obstacles. The map size of our indoor flight is $6m \times 6m \times2m$, and the resolution is $0.1m$. The maximum velocity and acceleration are set as $0.6m/s$ and $0.6m/{s^2}$. We employ the controller in \cite{c_2_19} to track the trajectory generated by our method. Fig. \ref{fig:flight} shows the snapshots of our experiment. The time of trajectory generation is $42ms$. The max velocity, acceleration, and jerk of trajectory are $0.53m/s$, $0.43m/{s^2}$ and $0.63m/{s^3}$, thus the whole trajectory is dynamically feasible and smooth. The minimum distance to the obstacle is $0.80m$ while no obstacle inflation is applied, which means the trajectory generated by our method is quite safe. We recommend readers to watch the attached video for more details.

\section{Conclusion}

In this paper, we propose a time-efficient search-based approach to generate safe, smooth and dynamically feasible trajectories for quadrotors. Benefiting from ESDF-based non-uniform search, our algorithm generates trajectories away from obstacles and has reasonable time-allocation. The highly strict convex hull property we derived makes the trajectory dynamically feasible and not too conservative. We have also enabled non-zero initial and termination states to be used so that algorithms can be applied in different scenarios. Extensions to this work may include developing informative heuristics for our search process and exploring the application of  the non-uniform rational B-spline curve in trajectory planning.

\addtolength{\textheight}{-12cm} 


\bibliographystyle{IEEEtran}
\bibliography{IEEEabrv,mybibfile}

\begin{thebibliography}{10}
\providecommand{\url}[1]{#1}
\csname url@rmstyle\endcsname
\providecommand{\newblock}{\relax}
\providecommand{\bibinfo}[2]{#2}
\providecommand\BIBentrySTDinterwordspacing{\spaceskip=0pt\relax}
\providecommand\BIBentryALTinterwordstretchfactor{4}
\providecommand\BIBentryALTinterwordspacing{\spaceskip=\fontdimen2\font plus
\BIBentryALTinterwordstretchfactor\fontdimen3\font minus
  \fontdimen4\font\relax}
\providecommand\BIBforeignlanguage[2]{{%
\expandafter\ifx\csname l@#1\endcsname\relax
\typeout{** WARNING: IEEEtran.bst: No hyphenation pattern has been}%
\typeout{** loaded for the language `#1'. Using the pattern for}%
\typeout{** the default language instead.}%
\else
\language=\csname l@#1\endcsname
\fi
#2}}

\bibitem{c_2_1}
D.~{Mellinger} and V.~{Kumar}, ``Minimum snap trajectory generation and control
  for quadrotors,'' in \emph{Proc. IEEE Int. Conf. Robotics and Automation},
  May 2011, pp. 2520--2525.

\bibitem{c_2_2}
E.~{Koyuncu} and G.~{Inalhan}, ``A probabilistic b-spline motion planning
  algorithm for unmanned helicopters flying in dense {3D} environments,'' in
  \emph{Proc. IEEE/RSJ Int. Conf. Intelligent Robots and Systems}, Sept. 2008,
  pp. 815--821.

\bibitem{c_2_4}
F.~{Gao}, W.~{Wu}, Y.~{Lin}, and S.~{Shen}, ``Online safe trajectory generation
  for quadrotors using fast marching method and bernstein basis polynomial,''
  in \emph{Proc. IEEE Int. Conf. Robotics and Automation (ICRA)}, May 2018, pp.
  344--351.

\bibitem{c_2_5}
C.~Richter, A.~Bry, and N.~Roy, ``Polynomial trajectory planning for aggressive
  quadrotor flight in dense indoor environments,'' in \emph{Robotics
  Research}.\hskip 1em plus 0.5em minus 0.4em\relax Springer, 2016, pp.
  649--666.

\bibitem{wang_2}
L.~Zhao, H.~Wang, and W.~Chen, ``Optimal trajectory planning for manipulators
  with flexible curved links,'' in \emph{International Conference on
  Intelligent Autonomous Systems}.\hskip 1em plus 0.5em minus 0.4em\relax
  Springer, 2016, pp. 1013--1025.

\bibitem{c_2_6}
S.~Karaman and E.~Frazzoli, ``Sampling-based algorithms for optimal motion
  planning,'' \emph{The international journal of robotics research}, vol.~30,
  no.~7, pp. 846--894, 2011.

\bibitem{wang_1}
Z.~Wei, W.~Chen, H.~Wang, and J.~Wang, ``Manipulator motion planning using
  flexible obstacle avoidance based on model learning,'' \emph{International
  Journal of Advanced Robotic Systems}, vol.~14, no.~3, p. 1729881417703930,
  2017.

\bibitem{c_2_7}
M.~Likhachev, G.~J. Gordon, and S.~Thrun, ``Ara*: Anytime a* with provable
  bounds on sub-optimality,'' in \emph{Advances in neural information
  processing systems}, 2004, pp. 767--774.

\bibitem{c_2_8}
Q.~H. {Do}, L.~{Han}, H.~{Tehrani Nik Nejad}, and S.~{Mita}, ``Safe path
  planning among multi obstacles,'' in \emph{Proc. IEEE Intelligent Vehicles
  Symp. (IV)}, June 2011, pp. 332--338.

\bibitem{c_2_9}
N.~{Ratliff}, M.~{Zucker}, J.~A. {Bagnell}, and S.~{Srinivasa}, ``Chomp:
  Gradient optimization techniques for efficient motion planning,'' in
  \emph{Proc. IEEE Int. Conf. Robotics and Automation}, May 2009, pp. 489--494.

\bibitem{c_2_10}
J.~Chen, T.~Liu, and S.~Shen, ``Online generation of collision-free
  trajectories for quadrotor flight in unknown cluttered environments,'' in
  \emph{2016 IEEE International Conference on Robotics and Automation
  (ICRA)}.\hskip 1em plus 0.5em minus 0.4em\relax IEEE, 2016, pp. 1476--1483.

\bibitem{c_2_11}
S.~{Liu}, M.~{Watterson}, K.~{Mohta}, K.~{Sun}, S.~{Bhattacharya}, C.~J.
  {Taylor}, and V.~{Kumar}, ``Planning dynamically feasible trajectories for
  quadrotors using safe flight corridors in {3-D} complex environments,''
  \emph{IEEE Robotics and Automation Letters}, vol.~2, no.~3, pp. 1688--1695,
  July 2017.

\bibitem{c_2_13}
W.~{Ding}, W.~{Gao}, K.~{Wang}, and S.~{Shen}, ``Trajectory replanning for
  quadrotors using kinodynamic search and elastic optimization,'' in
  \emph{Proc. IEEE Int. Conf. Robotics and Automation (ICRA)}, May 2018, pp.
  7595--7602.

\bibitem{c_2_12}
V.~{Usenko}, L.~{von Stumberg}, A.~{Pangercic}, and D.~{Cremers}, ``Real-time
  trajectory replanning for mavs using uniform b-splines and a {3D} circular
  buffer,'' in \emph{Proc. IEEE/RSJ Int. Conf. Intelligent Robots and Systems
  (IROS)}, Sept. 2017, pp. 215--222.

\bibitem{liu2017search}
S.~Liu, N.~Atanasov, K.~Mohta, and V.~Kumar, ``Search-based motion planning for
  quadrotors using linear quadratic minimum time control,'' in \emph{2017
  IEEE/RSJ International Conference on Intelligent Robots and Systems
  (IROS)}.\hskip 1em plus 0.5em minus 0.4em\relax IEEE, 2017, pp. 2872--2879.

\bibitem{c_2_14}
P.~{Fernbach}, S.~{Tonneau}, and M.~{Taïx}, ``Croc: Convex resolution of
  centroidal dynamics trajectories to provide a feasibility criterion for the
  multi contact planning problem,'' in \emph{Proc. IEEE/RSJ Int. Conf.
  Intelligent Robots and Systems (IROS)}, Oct. 2018, pp. 1--9.

\bibitem{c_2_16}
L.~Piegl and W.~Tiller, \emph{The NURBS book}.\hskip 1em plus 0.5em minus
  0.4em\relax Springer Science \& Business Media, 2012.

\bibitem{c_2_17}
K.~Qin, ``General matrix representations for b-splines,'' \emph{The Visual
  Computer}, vol.~16, no.~3, pp. 177--186, 2000.

\bibitem{wang_3}
H.~Wang, Y.~Lai, and W.~Chen, ``The time optimal trajectory planning with
  limitation of operating task for the tokamak inspecting manipulator,''
  \emph{Fusion Engineering and Design}, vol. 113, pp. 57--65, 2016.

\bibitem{c_2_18}
J.~{Bellingham}, A.~{Richards}, and J.~P. {How}, ``Receding horizon control of
  autonomous aerial vehicles,'' in \emph{Proc. American Control Conf. (IEEE
  Cat. No.CH37301)}, vol.~5, May 2002, pp. 3741--3746 vol.5.

\bibitem{c_2_19}
T.~{Lee}, M.~{Leok}, and N.~H. {McClamroch}, ``Geometric tracking control of a
  quadrotor {UAV} on se(3),'' in \emph{Proc. 49th IEEE Conf. Decision and
  Control (CDC)}, Dec. 2010, pp. 5420--5425.

\end{thebibliography}

\end{document}